# Improving Whole Slide Segmentation Through Visual Context - A Systematic Study

Korsuk Sirinukunwattana[1], Nasullah Khalid Alham[1,2], Clare Verrill[2], and Jens Rittscher[1]

[1] Institute of Biomedical Engineering, Department of Engineering Science, University of Oxford, Oxford
[2] Nuffield Department of Surgical Sciences and Oxford NIHR Biomedical Research Centre (BRC), University of Oxford, John Radcliffe Hospital, Oxford
{korsuk.sirinukunwattana|jens.rittscher}@eng.ox.ac.uk

**Abstract.** While challenging, the dense segmentation of histology images is a necessary first step to assess changes in tissue architecture and cellular morphology. Although specific convolutional neural network architectures have been applied with great success to the problem, few effectively incorporate visual context information from multiple scales. With this paper, we present a systematic comparison of different architectures to assess how including multi-scale information affects segmentation performance. A publicly available breast cancer and a locally collected prostate cancer datasets are being utilised for this study. The results support our hypothesis that visual context and scale plays a crucial role in histology image classification problems.

## 1 Introduction

Statistical learning approaches, primarily those embodied by deep learning, have demonstrated the potential for advancing our ability to extract information from histology images. The concept of end-to-end learning has been applied to predict cancer grade [1], genotype [2], and outcome [3] directly from the digitised haematoxylin and eosin (H&E) images. As opposed to summarising the vast amount of information in the form of a single number or category, we aim to capture potentially diagnostically relevant information and to support a more objective decision making process. Providing a dense segmentation of the entire image is a challenging and important first step towards achieving this goal.

Tissue architecture is characterised by an organ-specific hierarchical assembly of various components (e.g. stroma, epithelium, glands, blood vessels), their shape and topology. Progressing disease can severely disrupt this multi-scale organisation. Examples like those shown in Figure 1 illustrate how the increased amount of visual context improves the likelihood of correct identification. Classical medical imaging and computer vision research provides numerous examples on how information from multiple scales can be utilised. More recently, various deep learning approaches [4] have been introduced that effectively learn visual context directly from training data. With this paper we provide a more systematic comparison of these approaches and study how these effect the ability of differentiating between different tissue components. In addition, we introduce a computational model that utilises feature sharing across scales and learns dependencies between scales using long-short term memory (LSTM) unit [5].

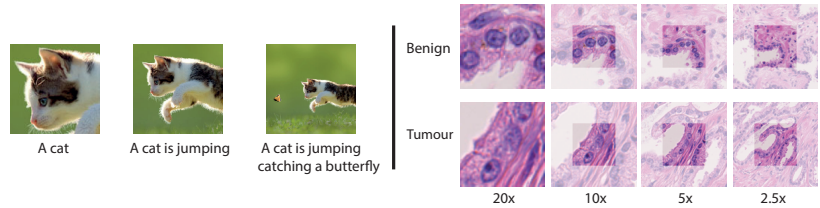

**Fig. 1. Visual context.** The different images of the scene containing a jumping cat effectively highlight that the correct interpretation of a scene depends on visual context. We content the accuracy of dense segmentation of histology images into different tissue types depends on our ability to make effective use of multiple scales.

An openly available collection of breast cancer samples [6] and a local collection of prostate cancer histology provide the necessary disease context. An overview of the relevant deep learning approaches is provided in Section 2. The set of architectures that are being used for a comparison and details of the datasets used in this study are being presented in Section 3. Our results in Section 4 give a strong indication that the modelling visual context impacts the quality of dense segmentation of histology images. While these results are extremely encouraging, we need to take shortcomings of the datasets into account. In our conclusions we outline what future studies are necessary to overcome the bias included in the present datasets.

## 2   Related Work

Two main approaches to medical image segmentations are semantic segmentation and patch-wise classification. For example, Ronneberger *et al.* [7] incorporate a dense prediction step in their U-Net convolutional neural network (CNN) architecture, which has been applied with great success to a range of biomedical applications. Zhang *et al.* [8] use a patch-based CNN approach to segment regions of infant MR brain images. In whole slide histology image segmentation, patch-based prediction models appear to dominate the landscape since the lack of comprehensive annotation ground-truth prohibits the use of semantic segmentation approach. Patch-based approaches have proven successful in various applications [9, 10]. To detect cancer metastases in breast atypical lymph nodes at a fine-grained level, Wang and colleagues [9] divide large whole slide images into small patches and employ a CNN to assign a prediction score to every patch. The final decision is aggregated from the micro predictions. Nonetheless, processing each patch independently does not take contextual information and long range spatial dependencies into account. To address this shortcoming, Moeskops *et al.* [11] extract patches of different sizes centred at the same pixel location. Each patch is processed on a separated branch of a CNN, yielding multiple-scale features which are then combined for the final prediction. Instead of extracting multiple patches at different scales, Kong *et al.* [12] use a CNN with a 2-dimensional long-short term (LSTM) architecture [4] to learn spatial dependencies of image patches and their neighbours. Incorporating multi-scale and contextual information into a patch-wise classification scheme is still an open problem. A systematic comparison of different network architectures is necessary to establish how visual context should be utilised in whole slide image segmentation.

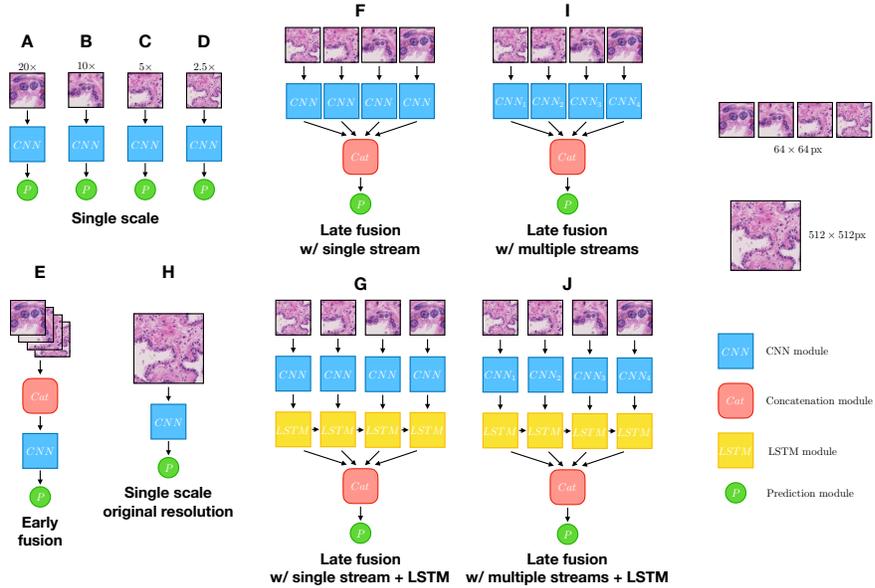

**Fig. 2. Used architectures.** Model complexity and run time are specified in Table 1.

## 3  Methods

***Comparative methods.*** The 10 different architectures that are being used in this study are presented in Figure 2. These can be categorised into three groups: 1) those that operates at a particular image resolution (A, B, C, D, and H), 2) those that fuse information at multiple resolutions before passing through a neural network (also known as early fusion approach, E), and 3) those that combine multi-scale output features from the networks before prediction (late fusion; F, G, I, and J). Two different approaches to late fusion are being considered. Architectures G and J apply an LSTM unit for integrating the multi-scale information, while methods F and I fuse the features as is. By using the same CNN for all scales (F & G) we test if it is beneficial to share features. In contrast, separate CNNs are being learnt for each spatial scale (I & J).

With the exception of model H the same CNN architecture is used to compare how the different approaches utilise the multi-scale context. It consists of 4 layers, where each layer starts with convolution of size $4 \times 4$, followed by a batch normalisation and rectify linear unit activation before downsampling. In method H, there are instead 8 of these convolutional layers. After these layers, the feature responses are fed to 2 fully connected layers, each with 512 hidden neural units. Dropout is employed after each fully connected layer. A linear classifer is used in the final layer of the network. The spatial dimensions of an input images in all methods is $64 \times 64$, except method H which uses large high-resolution images ($512 \times 512$) as input. The LSTM layer has a hidden state of size 512. See Supplementary Material for the detailed definition of each model.

The data are separated into 56% for training, 14% for validation, and 30% for testing at the patient level. The methods are trained using ADAM optimiser [13] with an initial learning rate of 0.0002. The training processed is stopped once the validation loss is no longer improving. Otherwise, the training is abort after 100 epochs.

*Datasets.* Two datasets are employed for quantitative evaluation. **Prostate:** the dataset consists of 4 tissue classes, including benign, lumen, stroma, and normal. Image patches are extracted from 28 whole slide images at 4 resolutions ($2.5\times$, $5\times$, $10\times$, and $20\times$). This implies that there are 4 images at an individual image location. There are no patches from the same whole slide image that appear in more than one of the training, validation, and test partitions. All annotations were provided by an expert prostate pathologist. In total, there are 41,442 patches at each scale before augmentation (lumen 8361, stroma 14547, benign 12,016, and tumour 6,518) **Breast:** this publically available dataset [6] consists of 4 tissue classes, namely normal, benign, in situ, and invasive. There are approximately 100 images in each classes. Training and test partitions are provided by the authors. Here, we extracted patches at 4 resolutions: $1.25\times$, $2.5\times$, $5\times$, and $10\times$. For each resolution, there are 27,060 patches before augmentation (normal 6,616, benign 7,050, in situ 8239, and invasive 5,155).

*Performance evaluation.* Since the segmentation problem is treated as a patch-based classification in this study, we consider an F1-measure for the performance evaluation. F1-measure is mathematically equivalent to the Dice index, a standard measure for segmentation accuracy. Due to the stochastic nature in the training process of the algorithms, we trained each approach 3 times, and in the evaluation, we use the average value of true positives, false positives, and false negatives across the 3 runs.

## 4   Results and Discussion

The results summarised in Table 1 provide some clear indication that including information from multiple different scales (E, F, G, I, and J) improves segmentation performance. When ranked with respect to performance, approaches that operate on a fixed resolution are clearly inferior. Rather than simply reporting out top performance with respect to each tissue category, we would like to highlight approaches that perform consistently well. A colour code is used in Table 1 to mark how each method relates to the top performer. In addition, we compute an accumulative rank.

**Table 1. Classification accuracy as measured by the F1-measure.** Bold indicates the best performance. Green, blue, yellow, and red colour codings indicate that the results are within 97.5%, 95%, 90%, and 85% of the best performance, respectively. This colour coding scheme can be used to rank the methods (bold = 1, green = 2, blue = 3, yellow = 4, red = 5, and no colour = 6). The overall ranking is summarised by the rank-sum. A total running time is measured on the test set of the prostate cancer data.

| Dataset | Class | Method | | | | | | | | | |
|---|---|---|---|---|---|---|---|---|---|---|---|
| | | A | B | C | D | E | F | G | H | I | J |
| Prostate | Lumen | 0.728 | 0.663 | 0.705 | 0.716 | 0.739 | 0.738 | 0.748 | 0.713 | 0.722 | **0.758** |
| | Stroma | 0.797 | 0.855 | 0.849 | 0.790 | 0.875 | 0.869 | 0.884 | **0.891** | 0.862 | 0.883 |
| | Benign | 0.508 | 0.646 | 0.712 | 0.717 | 0.734 | 0.745 | 0.766 | 0.763 | 0.765 | **0.782** |
| | Tumour | 0.562 | 0.653 | 0.629 | 0.579 | 0.699 | 0.687 | 0.728 | **0.746** | 0.674 | 0.712 |
| Breast | Normal | 0.501 | 0.468 | 0.523 | 0.513 | 0.509 | **0.603** | 0.573 | 0.252 | 0.241 | 0.323 |
| | Benign | 0.453 | 0.468 | 0.482 | 0.444 | 0.410 | 0.369 | 0.423 | **0.489** | 0.333 | 0.437 |
| | InSitu | 0.468 | 0.476 | 0.486 | 0.533 | **0.615** | 0.614 | 0.581 | 0.286 | 0.311 | 0.452 |
| | Invasive | 0.401 | 0.477 | 0.430 | 0.540 | 0.557 | 0.548 | 0.576 | 0.520 | 0.446 | **0.580** |
| Rank-sum (Prostate) | | 20 | 19 | 17 | 19 | 13 | 13 | 8 | 8 | 12 | 7 |
| Rank-sum (Breast) | | 22 | 21 | 19 | 18 | 16 | 13 | 14 | 18 | 24 | 18 |
| Total rank-sum | | 42 | 40 | 36 | 37 | 29 | 26 | 22 | 26 | 36 | 25 |
| No. of parameters | | 7.2M | 7.2M | 7.2M | 7.2M | 7.3M | 8.0M | 10.1M | 19.8M | 28.9M | 31.0M |
| Running time (s) | | 7.16 | 7.16 | 7.16 | 7.16 | 7.21 | 7.61 | 7.62 | 35.59 | 7.60 | 7.70 |

While model H yields the top performance for selected classes, it also performs rather poorly on others. Given that this model performs extremely well on detecting stroma in prostate tissue, one could argue that it specialises on capturing certain texture patters extremly well. When comparing models G and J we can make some interesting observations. On the given data sets model G performs consistently well in all of the tissue classes and has the lowest accumulative rank. Only considering the prostate samples model J is clearly the best. However, the performance of this model degrades on the breast cancer cases. Here, the interplay between model complexity and size of the data set needs to be taken into account. Later we discuss this issue in more detail. Overall, these results support our hypothesis that visual context and scale matters in histology image classification problems.

***Dataset size and makeup.*** It is crucial to mention that we observe a high degree of visual variation within each class in the breast cancer data. But each of these categories only contains a limited number of instances. This has two major consequences. When compared to the prostate experiments, all of the methods perform worse. More importantly, the breast cancer data set disadvantages more complex architectures. For example, consider method I and its counterpart with a significantly smaller number of parameters, method F. There is a dramatic drop in the performance of method I in relative to that of method F in most of the tissue classes. The same behaviour can be observed across a pair of methods J and G and a pair H and D in some of the tissue classes. This is why we need to interpret the results obtained on this breast data set with great caution.

***Feature integration.*** From Table 1, the models which utilise a LSTM unit (G and J) perform better than their counterparts with no LSTM (F and I) in most of the cases. Importantly, the LSTM unit also improves the resilience to noise. The direct comparison between models E and G shown in Figure 3 shows the percentage change in the F1-measure when we contaminate the images with noise. As one would expect, G is more resilient to noise as the percentage of reduction in the performance is consistently smaller than that of strategy E across tissue classes in both datasets.

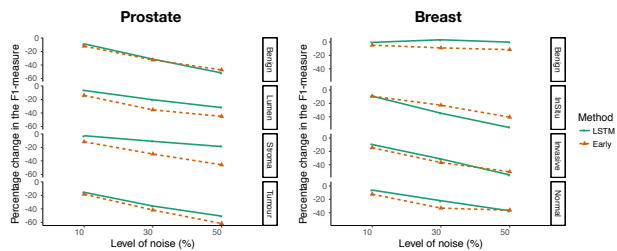

**Fig. 3. Resilience to noise.** The percentage change in the F1-measure at different noise levels of model E (in red) and model G (in green) is shown. The performance at the zero noise level is used as a reference. On each scale, an image is randomly replaced by a noisy image ($\sim \mathcal{N}(\cdot|\mu = 127, \sigma^2 = 1)$) with probability $p \in \{0.1, 0.3, 0.5\}$

***Computation efficiency vs accuracy.*** Especially when working on whole side images, computational efficiency needs to be taken into account. Here, memory usage and running time are important factors. In terms of the number of parameters, methods E, F and G and have a significantly lower number of parameters than H, I, and J. Based on the trend observed in the prostate dataset there is a possibility that when trained on more samples and longer methods I and J will yield an even better performance. On the other hand, methods E, F, G, I, and J run significantly faster than H (Table 1). In the medical context, the cost of running time has more weight than the cost of memory usage (number of parameters). As such, method H, which operates on the highest image resolution and full image dimensions, is considered very costly without offering significant improvement in the performance.

**Table 2. Effect of the order of image scales.** F1-measures of the method with a single stream of CNN and LSTM (G) subjected to different sequences of the image scale orders.

| Dataset | Class | Order | | | |
|---|---|---|---|---|---|
| | | Low → High | High → Low | Random | Bidirectional |
| Prostate | Lumen | 0.719 | 0.737 | 0.750 | **0.756** |
| | Stroma | 0.874 | 0.881 | 0.889 | **0.891** |
| | Benign | **0.790** | 0.779 | 0.787 | 0.776 |
| | Tumour | **0.750** | 0.706 | 0.724 | 0.734 |
| Breast | Normal | 0.573 | 0.588 | 0.590 | **0.609** |
| | Benign | **0.423** | 0.419 | 0.409 | 0.374 |
| | InSitu | **0.581** | 0.567 | 0.545 | 0.561 |
| | Invasive | **0.576** | 0.567 | 0.541 | 0.548 |

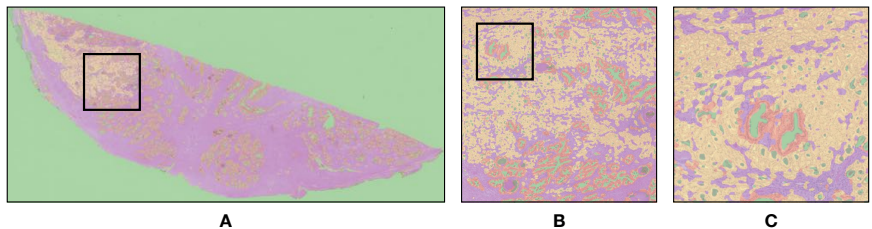

**Fig. 4. Segmentation example.** This whole slide segmentation was obtained with architecture G on one prostate cancer sample. Benign, tumour, lumen, and stroma regions are highlighted in orange, yellow, green, and purple. Note that the background white region is intentionally highlighted in green. Figures B and C correspond to the areas marked by rectangles in figure A and B.

***Sensitivity of method G to the order of image scales.*** To inspect whether the order of image scales affects the performance, we considered the following sequences of scale orders: 1) low to high, 2) high to low, 3) random ($5\times \rightarrow 2.5\times \rightarrow 20\times \rightarrow 10\times$ for the prostate and $2.5\times \rightarrow 1.25\times \rightarrow 10\times \rightarrow 5\times$ for the breast dataset), and 4) bidirectional (low $\leftrightarrow$ high). From Table 2, there is no strong difference between F1-measures produced by different experimental conditions. This implies that the performance of the method G does not depend on the order of the image scales.

## 5   Conclusions

To address the lack of comprehensive annotations we cast the segmentation problem as a patch-based classification rather than semantic segmentation task. In summary we conclude:

- **Visual context:** Our results support the claim that incorporating larger context produces superior results.
- **Feature integration:** LSTM units effectively capture the dependencies between different scales and generally improves performance. LSTMs are resilient to noise and not sensitive to the order of inputs.
- **Dataset design:** Small datasets typically do not represent the true variation of the data. Real clinical samples should be used for validation.

In addition, we have introduced a computationally efficient model (G) which performs well on various different tissue categories. Visual inspection of the segmentation results on whole slide images of this approach also looks highly encouraging (Figure 4). To overcome the problem of insufficient training data we aim at establish a standard dataset which includes data and annotation from multiple institutions. In addition to the manual annotation, immunohistochemistry staining will be considered to provide biological ground truth.

## 6   Acknowledgements

This research was funded by the National Institute for Health Research (NIHR) Oxford Biomedical Research Centre (BRC) and the EPSRC SeeBiByte

Programme Grant (EP/M013774/1). The views expressed are those of the authors and not necessarily those of the NHS, the NIHR or the Department of Health.

# Supplementary Material: Improving Whole Slide Segmentation Through Visual Context - A Systematic Study

Korsuk Sirinukunwattana[1], Nasullah Khalid Alham[1,2], Clare Verrill[2], and Jens Rittscher[1]

[1] Institute of Biomedical Engineering, Department of Engineering Science, University of Oxford, Oxford
[2] Nuffield Department of Surgical Sciences and Oxford NIHR Biomedical Research Centre (BRC), University of Oxford, John Radcliffe Hospital, Oxford
{korsuk.sirinukunwattana|jens.rittscher}@eng.ox.ac.uk

## 1 Model Definition

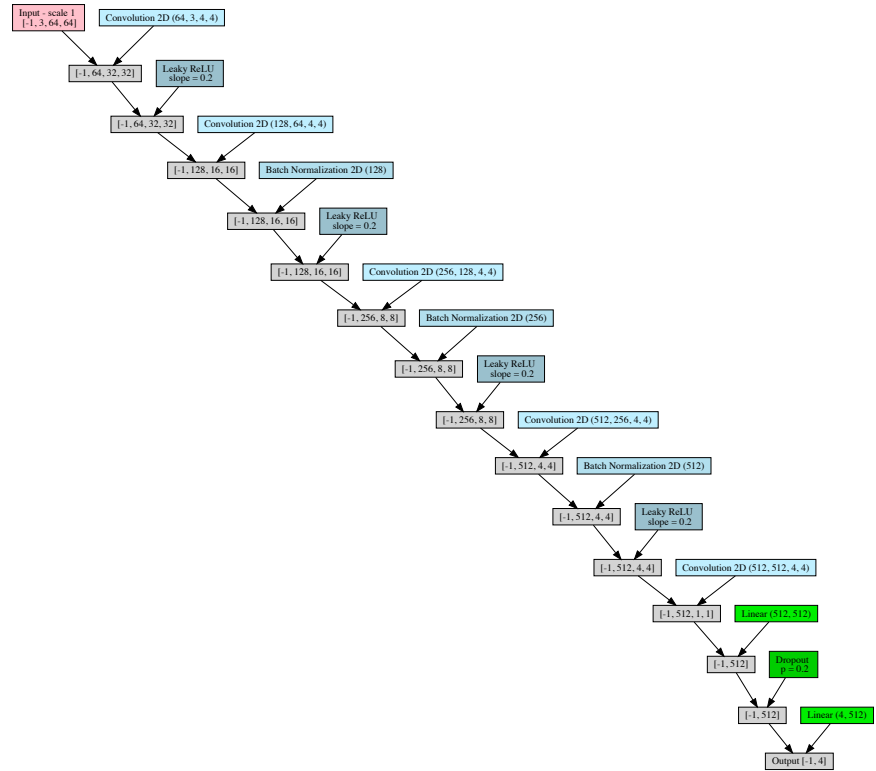

**Fig. 1.** Network definition of the model A, B, C, and D.

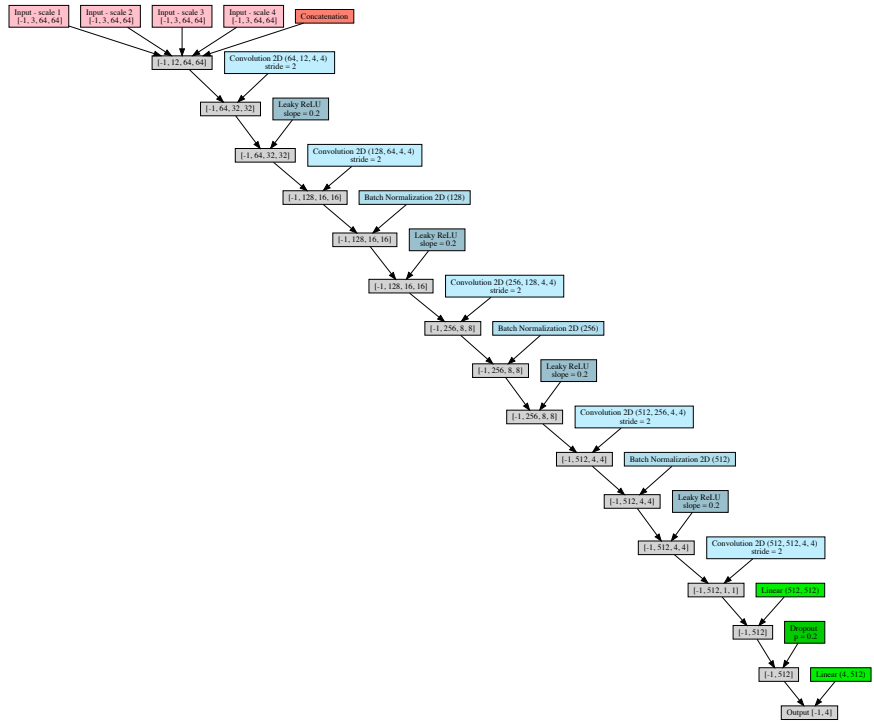

**Fig. 2.** Network definition of the model E.

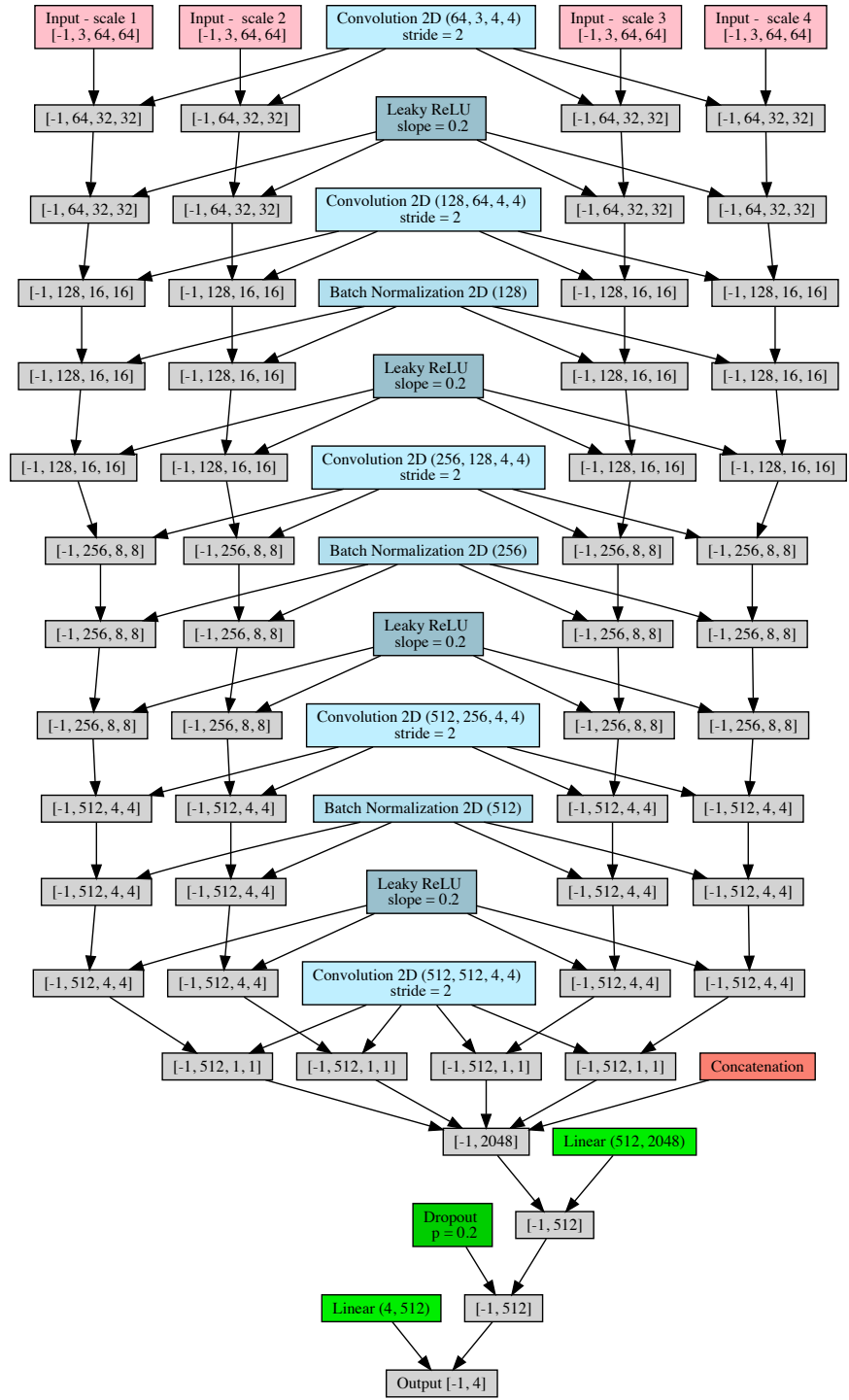

**Fig. 3.** Network definition of the model F.

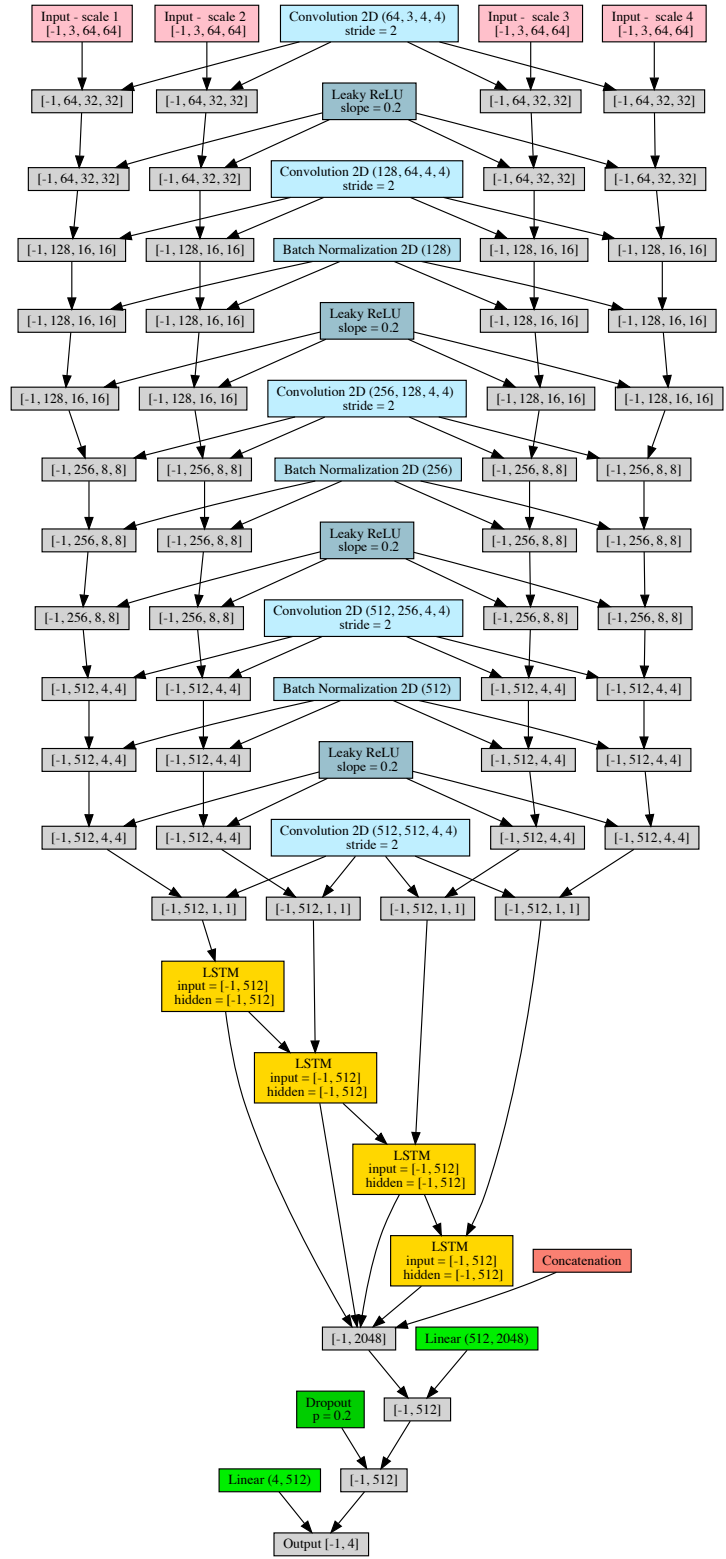

**Fig. 4.** Network definition of the model G.

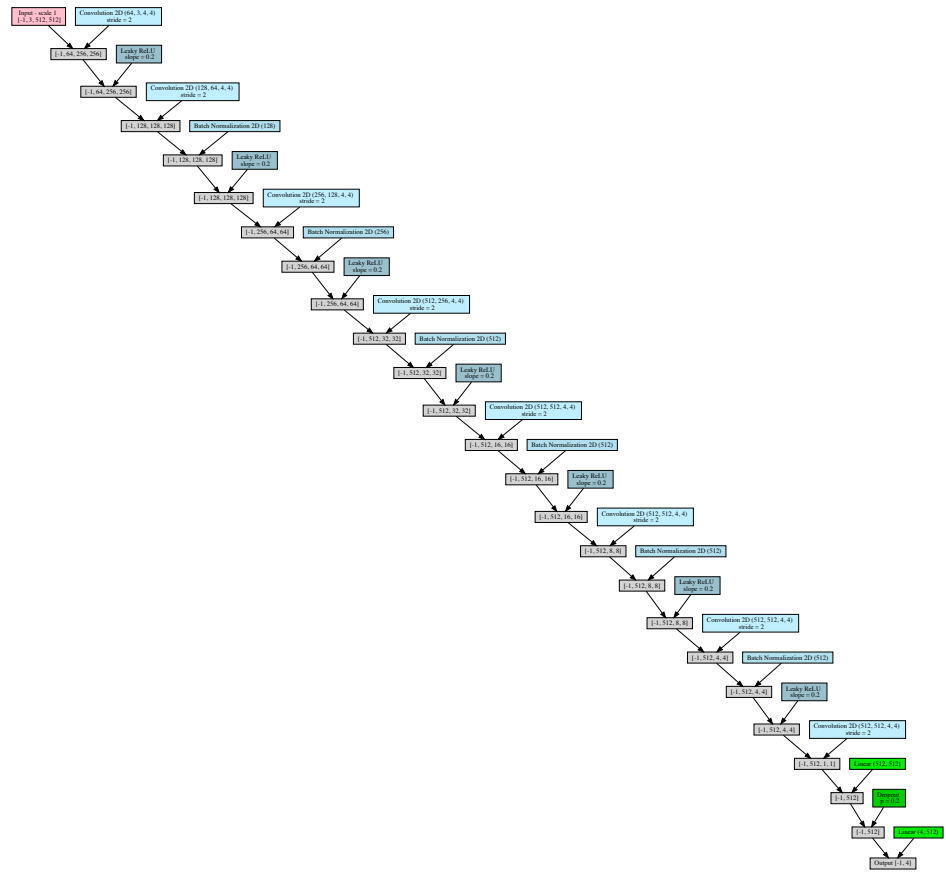

**Fig. 5.** Network definition of the model H.

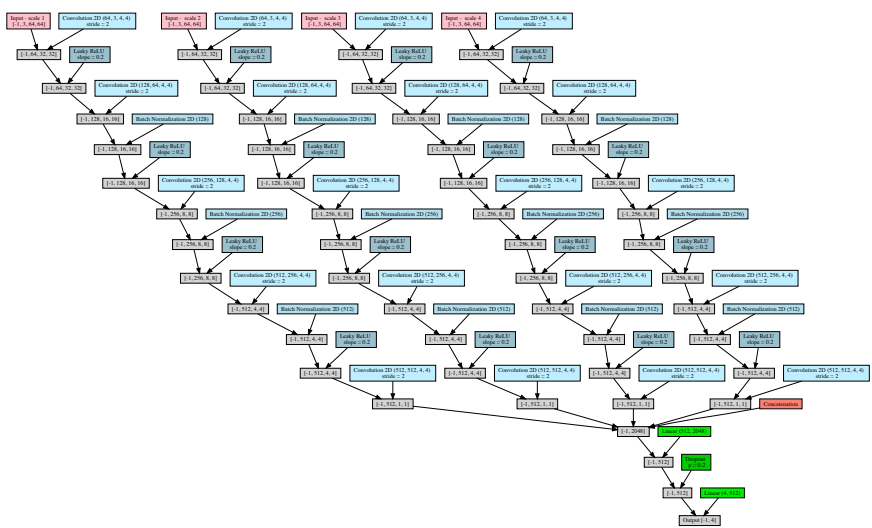

**Fig. 6.** Network definition of the model I.

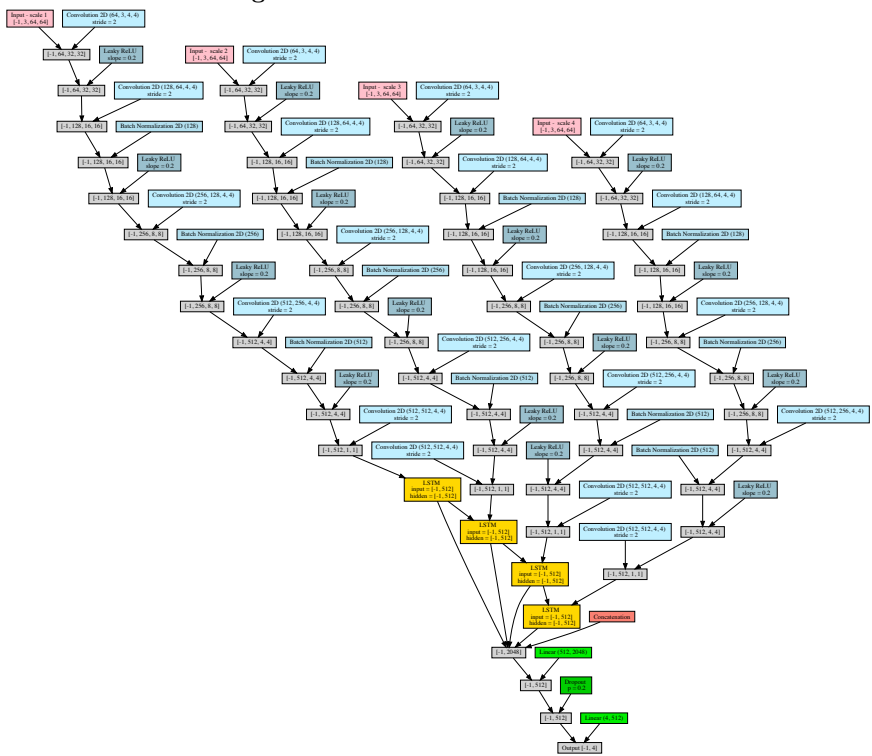

**Fig. 7.** Network definition of the model J.